\documentclass[conference]{IEEEtran}
\IEEEoverridecommandlockouts
\usepackage[T1]{fontenc}
\usepackage{amsmath,amsfonts}
\usepackage{array}
\usepackage{cite}
\usepackage[caption=false,font=scriptsize]{subfig}
\usepackage{textcomp}
\usepackage{stfloats}
\usepackage{url}
\usepackage{verbatim}
\usepackage{graphicx}
\usepackage{amssymb}
\usepackage{pifont}
\usepackage[]{todonotes} % disabled because it breaks tikzexternalize
\usepackage{hyperref}
\usepackage[capitalize]{cleveref}
\usepackage{comment}
\usepackage{multirow}
\usepackage{booktabs}
\usepackage{rotating}
\usepackage{makecell}
\usepackage{tabularx}
\usepackage{xcolor}
\usepackage[linesnumbered]{algorithm2e}
\RestyleAlgo{ruled}
\DontPrintSemicolon 

\usepackage{blindtext}
\usepackage{siunitx}
\AtBeginDocument{\RenewCommandCopy\qty\SI}
\usepackage{tikz}
\usetikzlibrary{arrows.meta}
\usetikzlibrary{arrows}
\usepackage{pgfplots}
\pgfplotsset{compat=newest}
\usetikzlibrary{external}
\usepackage{xcolor}
\definecolor{Bluelight}{HTML}{0065BD} % TUMBlue
\definecolor{Black}{HTML}{000000}
\definecolor{Blue}{HTML}{005293}
\definecolor{Bluestrong}{HTML}{003359}
\definecolor{Red}{HTML}{8C000F}
\definecolor{Grey}{HTML}{808080}
\definecolor{Greylight}{HTML}{CCCCCC}
\definecolor{Orange}{HTML}{E37222}
\definecolor{Green}{HTML}{A2AD00}
\definecolor{GreenCR}{HTML}{008000}
\definecolor{OrangeCR}{HTML}{f1b514}
\definecolor{visibleArea}{HTML}{81AF82}
\definecolor{occludedAreaObst}{HTML}{DC8282}
\definecolor{occludedAreaLanelet}{HTML}{DBBC82}

\newcommand{\visibleArea}{
  \tikz{
    \draw[fill=visibleArea] (0,0.0) rectangle (0.4,0.18);
  }
}

\newcommand{\occludedAreaObst}{
  \tikz{
    \draw[fill=occludedAreaObst] (0,0.0) rectangle (0.4,0.18);
  }
}

\newcommand{\occludedAreaLanelet}{
  \tikz{
    \draw[fill=occludedAreaLanelet] (0,0.0) rectangle (0.4,0.18);
  }
}

\newcommand{\bike}{
  \tikz{
    \node[inner sep=0pt] at (0,0) {\includegraphics[height=3.0mm]{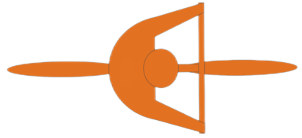}};
  }
}

\newcommand{\phantomBike}{
  \tikz{
    \node[inner sep=0pt] at (0,0) {\includegraphics[height=3.0mm]{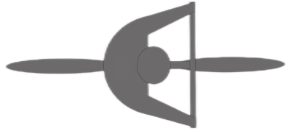}};
  }
}

\newcommand{\phantomPedestrian}{
  \tikz{
    \node[inner sep=0pt] at (0,0) {\includegraphics[height=3.0mm]{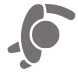}};
  }
}

\newcommand{\phantomPrediction}{
  \tikz{
    \fill[yellow,opacity=0.5] (0,0) circle (0.18);
    \fill[orange,opacity=0.5] (0,0) circle (0.1);
    \fill[red,opacity=0.5] (0,0) circle (0.04);
  }
}

\newcommand{\undetectedBike}{
  \tikz{
    \node[inner sep=0pt] at (0,0) {\includegraphics[height=3.0mm]{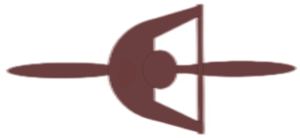}};
  }
}

\newcommand{\crashIcon}{
  \tikz{
    \node[inner sep=0pt] at (0,0) {\includegraphics[height=4.8mm]{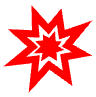}};
  }
}

\newcommand{\crashIconsmall}{
  \tikz{
    \node[inner sep=0pt] at (0,0) {\includegraphics[height=3.8mm]{figures/crashIcon.png}};
  }
}

\def\BibTeX{{\rm B\kern-.05em{\sc i\kern-.025em b}\kern-.08em
    T\kern-.1667em\lower.7ex\hbox{E}\kern-.125emX}}

\usepackage{fancyhdr}

\begin{document}

\title{\LARGE \bf Overcoming Blind Spots: Occlusion Considerations for Improved Autonomous Driving Safety}

\author{Korbinian Moller$^{1,*}$, Rainer Trauth$^{2,*}$, Johannes Betz$^{1}$

\thanks{$^{1}$ K. Moller, J. Betz are with the Professorship of Autonomous Vehicle Systems, TUM School of Engineering and Design, Technical University of Munich, 85748 Garching, Germany; Munich Institute of Robotics and Machine Intelligence (MIRMI). The authors gratefully acknowledge the financial support from the company Tier IV.}
\thanks{$^{2}$ The author is with the Institute of Automotive Technology, Technical University of Munich, 85748 Garching, Germany; Munich Institute of Robotics and Machine Intelligence (MIRMI). The author gratefully acknowledges the financial support from the Bavarian Research Foundation.}
\thanks{$^{*}$Shared first authorship.}}



\maketitle
\begin{abstract}
% Autonomous driving offers promising benefits but faces significant challenges in

Our work introduces a module for assessing the trajectory safety of autonomous vehicles in dynamic environments marked by high uncertainty. We focus on occluded areas and occluded traffic participants with limited information about surrounding obstacles. To address this problem, we propose a software module that handles blind spots (BS) created by static and dynamic obstacles in urban environments.
We identify potential occluded traffic participants, predict their movement, and assess the ego vehicle's trajectory using various criticality metrics. 
%Compared to other methods, our approach has a modular design and easy adaptability.
The method offers a straightforward and modular integration into motion planner algorithms. We present critical real-world scenarios to evaluate our module and apply our approach to a publicly available trajectory planning algorithm. Our results demonstrate that safe yet efficient driving with occluded road users can be achieved by incorporating safety assessments into the planning process. 
%However, it's important to note that our results have only been evaluated in simulations. 
The code used in this research is publicly available as open-source software and can be accessed at the following link: \url{https://github.com/TUM-AVS/Frenetix-Occlusion}.
\end{abstract}

\vspace{-0.8em}
%
%!TeX spellcheck = en_US
% !TeX root = ../main.tex

\section{Introduction}
\label{sec:introduction}

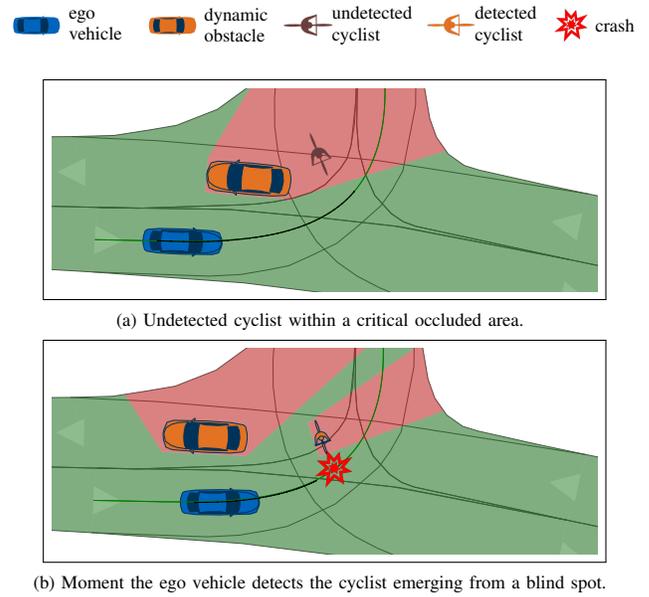
\begin{figure}[!t]
    \centering
    \hspace{1mm}
    %\vspace{-1cm}
    \begin{tikzpicture}[font=\scriptsize]
        % Ego vehicle
        \node[inner sep=0pt] at (0.9,0) {\includegraphics[height=2.5mm]{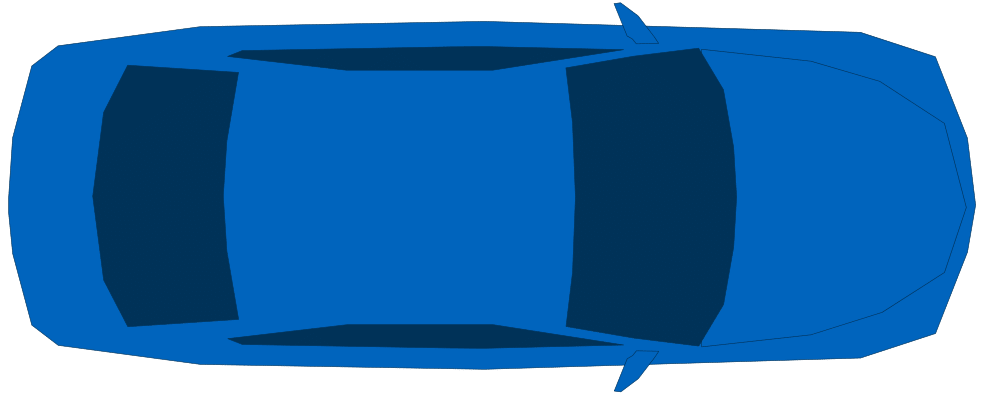}};
        \node[align=left, anchor=west] at (1.2,0) {ego \\ vehicle};
    
        % Obstacle
        \node[inner sep=0pt] at (2.7,0) {\includegraphics[height=2.5mm]{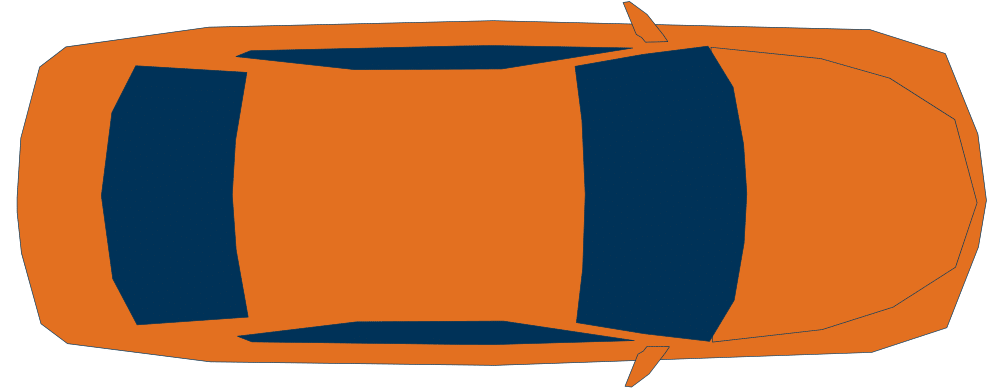}};
        \node[align=left, anchor=west] at (3.0,0) {dynamic \\ obstacle};

        % Phantom Bike
        \node[inner sep=0pt] at (4.5,0) {\undetectedBike};
        \node[align=left, anchor=west] at (4.7,0) {undetected \\ cyclist};

        % Bike
        \node[inner sep=0pt] at (6.4,0) {\bike};
        \node[align=left, anchor=west] at (6.6,0) {detected \\ cyclist};

        % Crash
        \node[inner sep=0pt] at (8.0,0) {\crashIcon};
        \node[align=left, anchor=west] at (8.2,0) {crash};
        
    \end{tikzpicture}
    %%%
    \vspace{-0.7em}
    \subfloat[][Undetected cyclist within a critical occluded area.]{\label{fig:introduction_PA} \fbox{\includegraphics[width=0.4\textwidth, trim={1.2cm 1.5cm 2.2cm 3.7cm},clip]{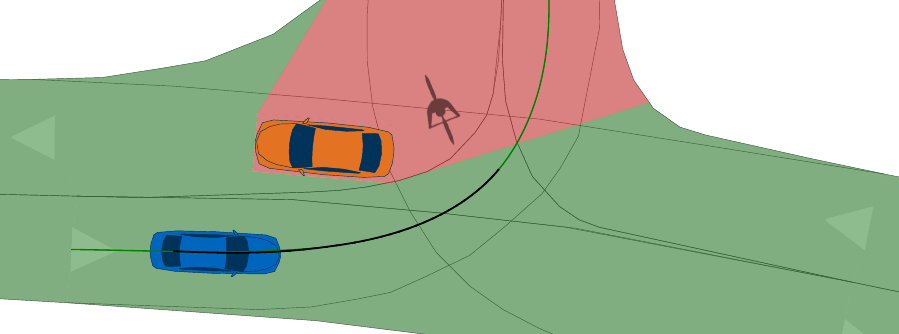}}} \\
    \subfloat[][Moment the ego vehicle detects the cyclist emerging from a blind spot.]{\label{fig:introduction_REAL} \fbox{\includegraphics[width=0.4\textwidth, trim={1.4cm 1.8cm 2.5cm 4.4cm},clip]{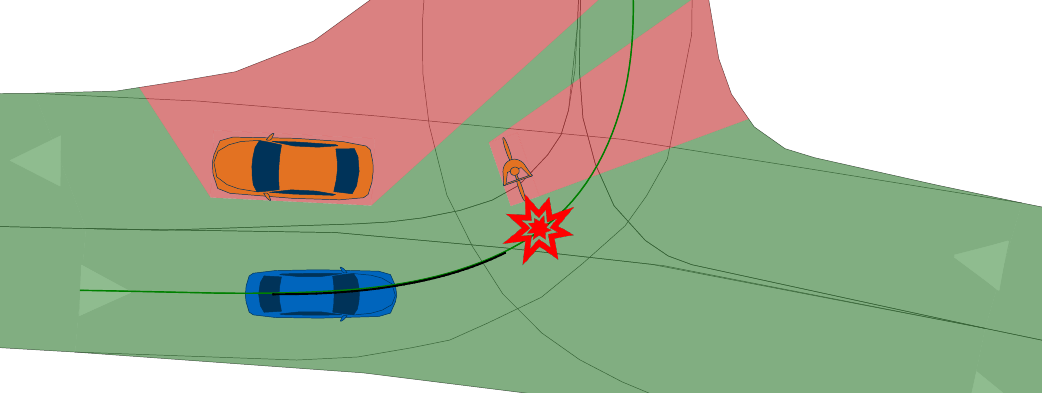}}} \\
    \caption{Exemplary visualization of a critical situation showcases an undetected cyclist emerging from a blind spot at an intersection, presenting a potential crash risk.}
    \label{fig:introduction}%
\end{figure}

Autonomous driving is emerging as a potential to revolutionize mobility, impacting our transportation systems and how we utilize cars~\cite{Bobisse.2019}. While promising significant benefits such as reducing traffic accidents, enhancing mobility for those unable to drive, and improving efficiency~\cite{Herrmann.2018}, the complete integration of autonomous vehicles (AVs) in dynamic environments remains a technical challenge. Especially in trajectory planning, a key challenge is the presence of occluded areas created by static and dynamic obstacles such as parked vehicles or other road users. Occluded areas may contain valuable information that can contribute to road safety. This particularly impacts undetected vulnerable road users (VRUs), e.g., pedestrians or cyclists (\cref{fig:introduction}). 
% The situation depicted in \cref{fig:introduction} showcases an undetected cyclist emerging from a blind spot at an intersection, presenting a potential crash risk.
Trajectory planning algorithms must, therefore, handle these uncertainties and unknowns arising from perception limitations and incorporate them into the planning process. %Furthermore, AVs must balance defensive driving and maintain efficient progress. 
%This is particularly important in complex scenarios like intersections, where AVs must account for potential hidden traffic participants in occluded areas. 
At the same time, AVs should not drive too defensively to avoid disrupting the traffic flow. \\
\indent This paper aims to tackle the challenges posed by occlusions in autonomous driving, focusing on enhancing safety while maintaining operational capability in the presence of blind spots (BS). To address this, we present the open-source module FRENETIX-Occlusion that evaluates trajectories by considering potential objects in occluded areas. This module uses a range of criticality metrics to conduct a comprehensive safety assessment of any given trajectory. The module supports trajectory selection by integrating this safety assessment into one's planning process to enhance autonomous driving in complex, dynamic environments. 
In summary, the new software module FRENETIX-Occlusion has three main contributions:
\begin{itemize}
    \item We present a method to \textbf{identify critical blind spots} in complex scenarios and model occluded \textbf{traffic participants as phantom agents (PA)}.
    \item We are able to predict various possible phantom agent movements. Subsequently, we \textbf{calculate criticality metrics} and make them available for further evaluation.
    \item The module is provided as an \textbf{open-source Python package} that can be integrated into existing trajectory planning algorithms.  
\end{itemize}
%!TeX spellcheck = en_US
% !TeX root = ../main.tex

\section{Related Work}
\label{sec:relatedwork}

Recent work has increasingly focused on overcoming challenges for AVs arising from occlusions. The approaches in this field of study can be categorized into several groups, each using different strategies to handle scenarios with occluded areas.

A key area of research focuses on assessing the maximum risk to pedestrians crossing streets to prevent collisions or minimize potential harm~\cite{Wang2023, Koc2021, TrauthOJITS}. This involves evaluating contextual information, such as visible vehicles or pedestrians, to determine the likelihood of unseen objects behind vehicles. Based on this assessment, a risk analysis is conducted and integrated into the trajectory planning algorithm. This approach can be generalized to other occluded road users, such as vehicles hidden behind curves \cite{Park2023}.
Another method employs particles to simulate the distribution of potentially occluded vehicles~\cite{Yu2020}. The particles are uniformly distributed along unobservable lanes and then advance in time at a steady speed. Subsequently, these particles are utilized to compute a collision risk, which can be used for trajectory evaluation. A similar approach calculates a visibility risk (VR), indicative of the potential collision risk with obstacles in occluded areas. It integrates the predicted VR into the cost function of a planning strategy~\cite{Wang.2020}.

The challenge of motion planning in environments with occluded obstacles has been addressed using Partially observable Markov decision processes (POMDPs)~\cite{Zhang2022, Wray2021}.% In these studies, motion planners based on POMDPs are utilized to navigate complex scenarios with occluded areas. 
The algorithms are often enhanced by incorporating contextual appearance probabilities. Hierarchical decision-making methods have been developed, particularly useful in specific scenarios like intersections~\cite{Wray2021}. These methods consist of a dual-framework approach: a higher-level candidate path selector for preliminary decision-making and a lower-level POMDP planner for detailed vehicle navigation. In~\cite{Zhang2022}, the authors have implemented phantom vehicles and pedestrians to evaluate potential risks in challenging scenarios efficiently. Following POMDPs, game-theoretic methods provide an alternative approach, facilitating less conservative motion planning by considering interactions between traffic participants~\cite{ZhangZixu.2021}. While these strategies offer dynamic solutions, ensuring safety is more complex, particularly when involving interactive traffic participants.

In the work of Koschi et al.~\cite{Koschi2021}, a predictive approach that accounts for visible and potentially occluded traffic participants is presented. Their method involves applying formalized traffic rules and motion models to conduct a reachability analysis, predicting the possible locations and speeds of vehicles, pedestrians, and cyclists. The set-based prediction method offers a versatile solution capable of generalizing across various traffic situations with occluded areas~\cite{Orzechowski2018}. Another notable contribution is a study on autonomous valet parking in limited-visibility environments~\cite{Lee.2021}. This method utilizes reachable set estimation to account for obstacles and enable safe vehicle movements, considering vehicle motion constraints and leveraging sensor data to estimate the space around the vehicle. Its application aims to facilitate collision-free parking maneuvers. The reachable set approach can be extended to ensure collision-free driving in all traffic situations~\cite{Nager2019}. In~\cite{Sanchez.2022}, the authors address the issue of occlusion-aware motion planning by considering all possible unseen traffic participants. They achieve this by utilizing reachable sets to calculate the future positions of these participants and eliminating implausible obstacle states based on previous observations.

Finally, enhanced safety in autonomous driving can also be achieved by expanding the visible area through lateral position adjustments~\cite{Gilhuly2022, Narksri2022}. Central to this method is a cost function that assesses the visibility of occluded regions. The metric is integrated into the motion planning algorithm, fostering the generation of trajectories prioritizing a broader field of vision.

%!TeX spellcheck = en_US
% !TeX root = ../main.tex

\section{Methodology}
\label{sec:method}
\begin{figure*}[!ht]
    \centering
    \input{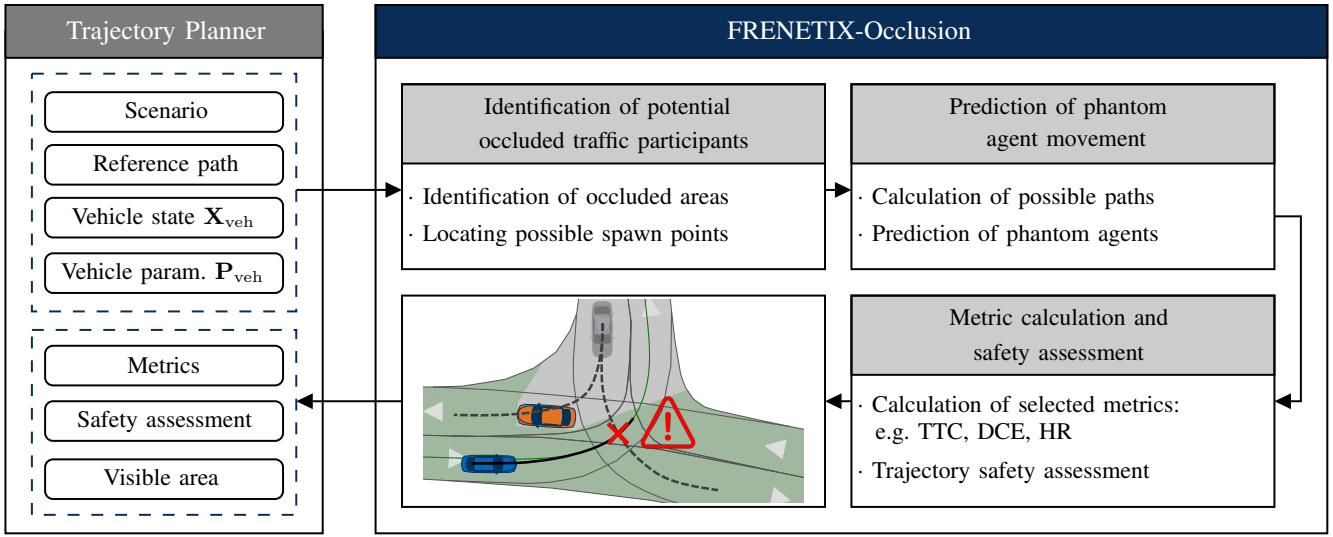}
    \caption{FRENETIX-Occlusion software module framework with required inputs and provided outputs.}
    \label{fig:overview}
\end{figure*}

To overcome the shortcomings presented in the state of the art, we present the methodology behind the FRENETIX-Occlusion module to enhance motion planning algorithms for AVs with occlusion awareness. % We augment existing methodologies to forge an open-source trajectory safety assessment tool tailored for autonomous driving in occluded scenarios.
FRENETIX-Occlusion identifies potential occluded traffic participants within critical occluded areas, predicts possible movements, and assesses the AV's trajectory to ensure safety. An overview of our framework is depicted in \cref{fig:overview}.
%The module requires environmental information, which the trajectory planner must provide. pinpoints occluded regions, and determines locations where these participants may emerge. The next step involves modeling phantom agents (PAs) and calculating their potential trajectories to predict future positions (\cref{subsec:Prediction}). The final steps are the computation of critical safety metrics for the proposed trajectories (\cref{subsec:Metric}) and a safety check that returns whether a trajectory is valid or not (\cref{subsec:SafetyCheck}). %Detailed explanations of each component will be provided in the following sections. 

\subsection{Required Semantic Information and Framework}
\label{subsec:InputOutput}
FRENETIX-Occlusion is designed to operate with trajectory planners that employ a curvilinear coordinate system to a reference path ($\Gamma$), with the lanelet-based environment ($\mathcal{L}$). % The scenario encapsulating these details and the reference path must be available. 
The vehicle state vector $\mathbf{X}_{\mathrm{veh}}$ is required, including the vehicle's $x$, $y$ coordinates, longitudinal $s$ and lateral $d$ curvilinear coordinates, orientation $\theta$ and velocity $v$. Vehicle parameters $\mathbf{P}_{\mathrm{veh}}$ such as length $\text{l}_\mathrm{veh}$, width $\text{w}_\mathrm{veh}$, wheelbase $\text{b}_\mathrm{veh}$, and sensor specifications (e.g., sensor range  $\text{r}_\mathrm{veh}$) are also necessary. %
\begin{equation*}
    \mathbf{X}_{\mathrm{veh}} = \begin{bmatrix}
    x, y, s, d, \theta, v
\end{bmatrix}^\intercal, 
\; 
\mathbf{P}_{\mathrm{veh}} = \begin{bmatrix}
\text{l}_\mathrm{veh}, 
\text{w}_\mathrm{veh},
\text{b}_\mathrm{veh},
\text{r}_\mathrm{veh}
\end{bmatrix}^\intercal
\end{equation*}
Upon completion of the calculation, the metrics and the result of the safety assessment are available as outputs. Optionally, the visible area calculated by the sensor model can be returned.
\subsection{Identification of Potential Occluded Traffic Participants}
\label{subsec:Identification}
Information about visible and occluded regions is required to place phantom agents (PAs), which we use to model unseen traffic participants. Once these areas are determined, potential spawn points (SP) can be established.

\begin{figure}[!ht]
    \centering
    \hspace{1mm}
    %\vspace{-1cm}
    \begin{tikzpicture}[font=\scriptsize]
        % Ego vehicle
        \node[inner sep=0pt] at (0.7,0) {\includegraphics[height=2.5mm]{figures/ego.png}};
        \node[align=left, anchor=west] at (1.0,0) {ego \\ vehicle};
    
        % Obstacle
        \node[inner sep=0pt] at (2.4,0) {\includegraphics[height=2.5mm]{figures/dyn_obstacle.png}};
        \node[align=left, anchor=west] at (2.7,0) {dynamic \\ obstacle};
    
         % Sensor radius
        \draw[thick, black, dashed, line width=1pt] (3.9,0) -- (4.3,0);
        \node[align=left, anchor=west] at (4.3, 0) {sensor \\ radius};

        % visible area
        \node[inner sep=0pt] at (5.6,0) {\visibleArea};
        \node[align=left, anchor=west] at (5.8,0) {$\mathcal{A}_\mathrm{v}$};

        % occluded area obstacle
        \node[inner sep=0pt] at (6.7,0) {\occludedAreaObst};
        \node[align=left, anchor=west] at (6.9,0) {$\mathcal{A}_\mathrm{o}$};

        % occluded area lanelet
        \node[inner sep=0pt] at (7.8,0) {\occludedAreaLanelet};
        \node[align=left, anchor=west] at (8.0,0) {$\mathcal{A}_\mathrm{b}$};
        
    \end{tikzpicture}
    \includegraphics[width=0.90\linewidth, trim={12cm 6cm 12cm 6cm},clip]{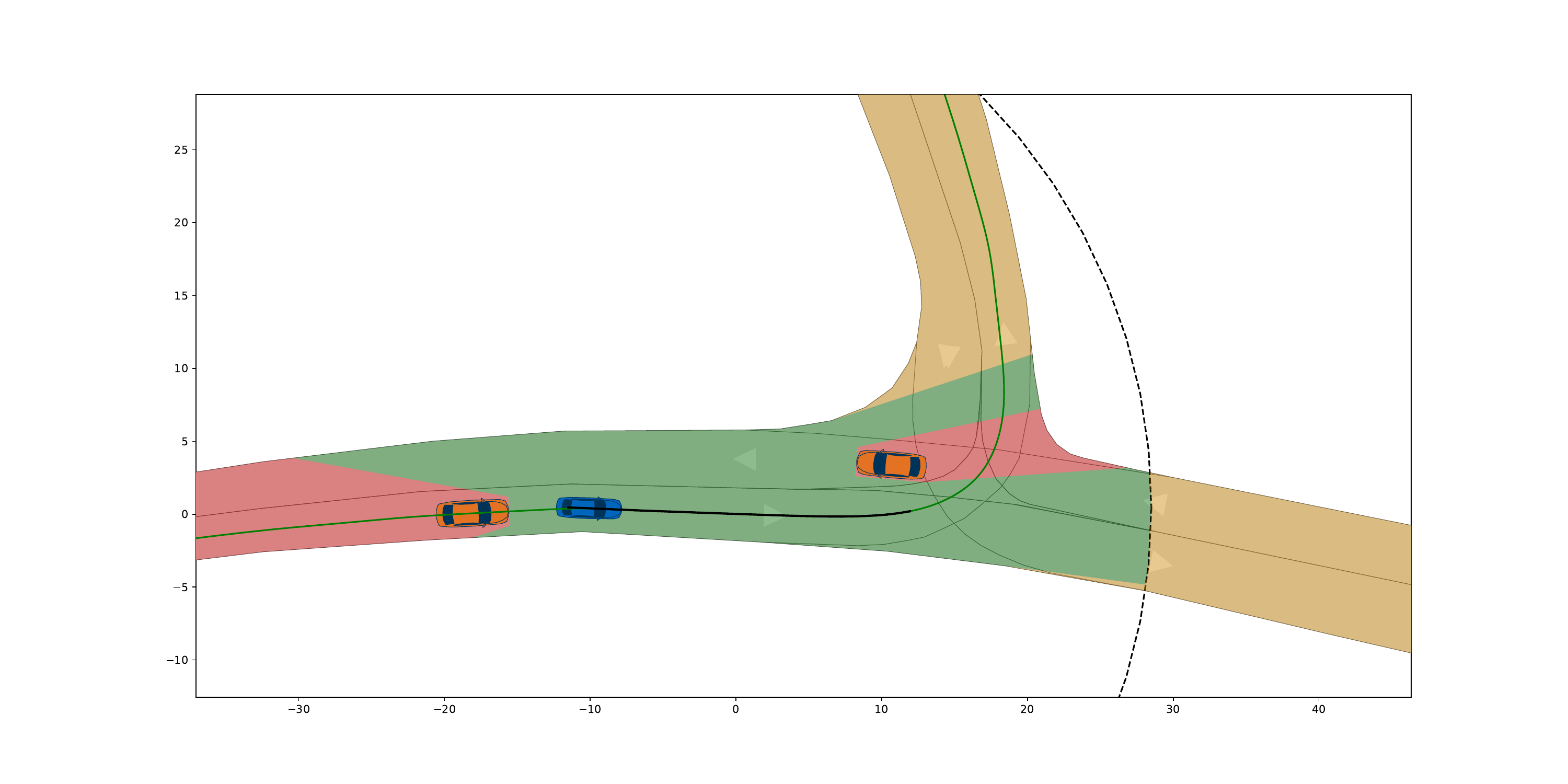}
    \caption{Illustration of visible and occluded areas. The visible area $\mathcal{A}_\mathrm{v}$ represents the region within the sensor range that is not obscured by any obstacles. The occluded areas indicate zones occluded by dynamic obstacles $\mathcal{A}_\mathrm{o}$ and boundaries $\mathcal{A}_\mathrm{b}$.}
    \label{fig:sensormodel}
\end{figure}

\textbf{Visible and occluded areas:} Our module includes a sensor model that uses geometric and semantic data to identify visible and occluded regions. Each area is stored as a polygon, accurately representing the environmental layout. The sensor model updates at every timestep $k$ to ensure continuous precision and reliability. 
%We define the visible area $\mathcal{A}_\mathrm{v}$ according to \cite{TrauthOJITS, Sanchez.2022} as the region obtained by removing the occluded area behind boundaries $\mathcal{A}_\mathrm{b}$ and obstacles $\mathcal{A}_\mathrm{o}$ from the intersection of the sensor radius area $\mathcal{A}_\mathrm{sr}$ and the available area of the lanelet network $\mathcal{A}_\mathrm{ln}$, as represented in \cref{eq:sensormodel}: %
We define the visible area $\mathcal{A}_\mathrm{v}$, as per \cite{TrauthOJITS, Sanchez.2022}, to be the region derived from the intersection of the sensor radius area $\mathcal{A}_\mathrm{r}$ and the lanelet network's available area $\mathcal{A}_\mathcal{L}$. This definition involves removing the occluded area behind boundaries $\mathcal{A}_\mathrm{b}$ (e.g. a house) and other traffic obstacles $\mathcal{A}_\mathrm{o}$ (e.g. a car), as illustrated in \cref{eq:sensormodel}.
\begin{equation}
\label{eq:sensormodel}
\mathcal{A}_\mathrm{v} = (\mathcal{A}_\mathrm{r} \cap \mathcal{A}_\mathcal{L}) \setminus \underbrace{(\mathcal{A}_\mathrm{b} \cup \mathcal{A}_\mathrm{o})}_{\mathcal{A}_\mathrm{occ}}
\end{equation}
The described areas are visualized in \cref{fig:sensormodel}. %In this depiction, the resulting visible area is colored green, the occluded areas based on lanelet geometry are shown in orange, and the areas occluded behind obstacles are marked in red. The sensor radius is outlined in black, and 
A conservative assumption is employed to calculate the visible area where the sensor's detection capability does not extend beyond the road boundaries. This scenario is commonly encountered in urban or residential areas with buildings lining the streets.

\noindent After calculating the visible and occluded areas, we identify occluded regions that constitute critical blind spots. For an area to be considered an acute occlusion, it must be of sufficient size to geometrically accommodate a PA. To identify potential SP, three distinct types of blind spots are differentiated. These spots are generated by various causes (\cref{tab:blind_spots}).
\begin{table}[!htb]
\centering
\caption{Critical blind spots (BS).}
\begin{tabularx}{0.72\linewidth}{p{2.0cm} p{1.9cm} p{1.9cm} }
    \toprule
    \textbf{Cause of BS }& \textbf{relevant at} & \textbf{PA type} \\
    \midrule
    & $\cdot$ left turn & \\
    Static obstacle & $\cdot$ right turn & $\cdot$ pedestrian \\
    & $\cdot$ straight & \\
    \midrule
    \multirow{2}{*}{Lane geometry} & $\cdot$ left turn & \multirow{2}{*}{$\cdot$ pedestrian} \\
    & $\cdot$ right turn &  \\
    \midrule
    \multirow{2}{*}{Dynamic obstacle} & $\cdot$ left turn & $\cdot$ bicycle \\
    & $\cdot$ straight & $\cdot$ vehicle \\
    \bottomrule
    \end{tabularx}
\label{tab:blind_spots}
\end{table}

\textbf{Blind spots caused by static obstacles:} %
Areas behind parked vehicles are particularly significant, as they pose risks for crossing pedestrians \cite{Verkehrsunfaelle2022}. To identify SP behind static obstacles, we define the visible static obstacles as $\mathcal{O}_{\mathrm{v, stat}}$ and sort them by their Euclidean distance to the ego AV. An obstacle's minimum and maximum extent is computed if the distance to the AV is within a specified threshold. Subsequently, lines perpendicular to the reference path, denoted as ${L}_{\mathrm{perp}}$, are examined to identify where they intersect with the boundary between $\mathcal{A}_{\mathrm{v}}$ and $\mathcal{A}_{\mathrm{occ}}$. These intersection points, labeled as $\mathcal{P}_{\mathrm{inter}}^\mathrm{stat}$ serve as SP candidates $\mathcal{C}_\mathrm{SP}^\mathrm{stat}$. 

\textbf{Blind spots caused by lane geometry:} %
Additionally, areas obscured by curves are critical, especially in right-turn scenarios where visibility is significantly limited. In many cases, pedestrians crossing behind curves can only be seen at close range, increasing the risk of collision. Our approach involves calculating potential pedestrian SPs in such scenarios to address this. We analyze the intersection of the ego vehicle's reference path $\Gamma$, with the occluded areas $\mathcal{A}_\mathrm{occ}$ identified by the sensor model. 
\begin{equation}
    \mathcal{P}_\mathrm{inter}^\mathcal{L} = \Gamma \; \cap \; \mathcal{A}_{\mathrm{occ}}
\end{equation}
From these intersections $\mathcal{P}_\mathrm{inter}^\mathcal{L}$, we derive SP candidates $\mathcal{C}_\mathrm{SP}^\mathcal{L}$. \\
In both static and lane-geometry-based cases, we select the final spawn positions $\mathcal{P}_\mathrm{SP}$ from the candidates $\mathcal{C}_\mathrm{SP}$ identified in each case, applying the following criteria:

\begin{itemize}
    \item $\mathcal{P}_\mathrm{SP}$ is a candidate $\mathcal{C}_{\mathrm{SP}}$
    \item $\mathcal{P}_\mathrm{SP}$ is within the lanelet network $\mathcal{A}_{\mathcal{L}}$
    \item $\mathcal{P}_\mathrm{SP}$ is not intersecting with any visible obstacles $\mathcal{O}_\mathrm{v}$
\end{itemize}
In summary, the identification of valid SPs behind static obstacles and turns is represented as:
\begin{equation}
    \mathcal{P}_{\mathrm{SP}}^{\mathrm{stat} \; \cup \; \mathcal{L}} = \{ p \;|\; p \in \mathcal{C}_{\mathrm{SP}} \; \land \; p \in \mathcal{A}_{\mathcal{L}} \; \land \; p \not\in \mathcal{O}_\mathrm{v} \; \}
\end{equation}

\textbf{Blind spots caused by dynamic obstacles:} %
Furthermore, occlusions caused by dynamic obstacles constitute a significant concern. These obstacles often conceal other road users, including vehicles and, more critically, cyclists. In intersection scenarios, such occlusions can lead to hazardous situations. The module identifies SPs behind dynamic obstacles within a specified proximity to the ego vehicle. %
We denote the set of dynamic obstacles within the visible range as $\mathcal{O}_{\mathrm{v, dyn}}$. For each obstacle $o \in \mathcal{O}_{\mathrm{v, dyn}}$, we compute the relative curvilinear coordinates $(s_\mathrm{o}, d_\mathrm{o})$ to the ego vehicle, considering only those within a predefined distance $d_{\mathrm{max}}$. The occluded area $\mathcal{A}_{\mathrm{occ}}^\mathrm{o}$ behind an obstacle is then determined. Subsequently, it is assessed whether the minimal required PA area $\mathcal{A}{\mathrm{occ}}$ can be placed around the centroid of $\mathcal{A}{\mathrm{occ}}^\mathrm{o}$. SP candidates $\mathcal{C}_{\mathrm{SP}}^{\mathrm{dyn}}$ are derived from these identified locations, ensuring they are plausible points of PA appearance. The final output comprises potential PA spawn locations:
\begin{equation}
\mathcal{P}_{\mathrm{SP}}^\mathrm{dyn} = \{ p \;|\; p \in \mathcal{C}_{\mathrm{SP}}^\mathrm{dyn} \land p \notin \mathcal{O}_{\mathrm{v}} \}
\end{equation}

\noindent The aggregated set, represented as: 
\begin{equation}
\mathcal{P}_{\mathrm{SP}} = \mathcal{P}_{\mathrm{SP}}^{\mathrm{stat} \; \cup \; \mathcal{L}} \; \cup \; \mathcal{P}_\mathrm{SP}^\mathrm{dyn}  
\end{equation}
encapsulates the complete spectrum of potential SPs and is forwarded to the prediction of our module.
\subsection{Prediction of Phantom Agent Movement}
\label{subsec:Prediction}
Our module generates potential paths initialized at the identified spawn points $\mathcal{P}_\mathrm{SP}$, considering all movement possibilities, such as straight, right, or left turns. Trajectory predictions are calculated for each path, projecting the future positions of the PAs using a constant velocity model, as presented in Algorithm~\ref{algo:PAPrediction}. The specific type of each PA is important in shaping the trajectory characteristics. For vehicles and cyclists, trajectories are aligned with traffic rules, adhering to lane discipline and moving at relatable velocities:
\begin{equation}
\begin{split}
{\xi}_{\mathrm{veh}}, \; {\xi}_{\mathrm{cyc}} \in \mathcal{L}_{\mathrm{allowed}} \subset \mathcal{L}, \\
{v}_{\mathrm{veh}}(t), \; v_{\mathrm{cyc}}(t) \in v_{\mathrm{allowed}}, 
\end{split}
\end{equation}
where ${\xi}_{\mathrm{veh}}$ and ${\xi}_{\mathrm{cyc}}$ represent the predicted trajectories of vehicles and cyclists. $\mathcal{L}_{\mathrm{allowed}}$ denotes the set of permissible lanelets as a subset of the entire lanelet network $\mathcal{L}$. \\
For pedestrians, the trajectory ${\xi}_{\mathrm{ped}}$ is a direct path across the road, orthogonal to the reference path.
\begin{equation}
   \xi_\mathrm{ped} \perp \Gamma
\end{equation}
The module uses these predicted PA trajectories to calculate the criticality metrics.

\begin{algorithm}[!t]
\SetAlgoLined
\SetKwInOut{Input}{Input}\SetKwInOut{Output}{Output}
\SetKwFunction{FRP}{findPossibleRoutes}
\SetKwFunction{SIV}{setInitialVelocity}
\SetKwFunction{GIO}{getInitialOrientation}
\SetKwFunction{FCL}{findCurrentLanelet}
\SetKwFunction{FIS}{InitialState}
\SetKwFunction{FCT}{createPrediction}
\SetKwFunction{IV}{isValid}

\Input{Spawn Point $\mathcal{P}_{\mathrm{SP, PA}}$, Agent Parameters $\mathbf{P}_{\mathrm{PA}}$}
\Output{Trajectory Prediction $\xi_\mathrm{PA}$}

\BlankLine
$\mathcal{L}_\mathrm{PA} \leftarrow$ \FCL{$\mathcal{P}_{\mathrm{SP, PA}}$}\;
$\mathcal{R}_\mathrm{PA} \leftarrow$ \FRP{$\mathcal{L_\mathrm{PA}}$}\;
$v_\mathrm{PA} \leftarrow$ \SIV{$\mathcal{L_\mathrm{PA}}, \mathbf{P}_\mathrm{PA}$}\;
$\Theta_\mathrm{PA} \leftarrow$ \GIO{$\mathcal{L_\mathrm{PA}}, \mathbf{P}_\mathrm{PA}$}\;
$\mathbf{X}_\mathrm{PA} \leftarrow$ \FIS{$\mathcal{P}_\mathrm{SP, PA}, v_\mathrm{PA}, \Theta_\mathrm{PA}$} with $v_\mathrm{PA}$ and $\Theta_\mathrm{PA}$ on $\mathcal{L}_\mathrm{PA}$\;

\BlankLine

\ForEach{$r \in \mathcal{R}_\mathrm{PA}$ }{
    \If{\IV{$r$}}
    {
        $\xi_\mathrm{PA}^{r} \leftarrow$ \FCT{$r, \mathbf{X}_\mathrm{PA}$} 
    }
}
$\xi_\mathrm{PA} \leftarrow \langle \xi_\mathrm{PA}^{r1}, \hdots,  \xi_\mathrm{PA}^{rN} \rangle$

\Return $\xi_\mathrm{PA}$ PA trajectories with different routes \;

\caption{Predict PA movement \label{algo:PAPrediction}}
\end{algorithm}

\subsection{Occlusion-criticality Measurement}
\label{subsec:Metric}
We employ specific metrics that allow real-time evaluation of trajectories rather than post hoc analysis. This enables immediate assessment and adjustment of the trajectory in response to dynamic conditions and the defined occlusions. \cref{tab:Metrics} shows the implemented criticality metrics. 
\begin{table}[!ht]
\centering
\caption{List of implemented metrics from \cite{Lin.2023}.}
\begin{tabularx}{0.76\linewidth}{l l c}
    \toprule
    Acronym & Measure & Source \\
    \midrule
    TTC & Time-to-collision & \cite{Sontges.2018} \\
    WTTC & Worst-time-to-collision & \cite{Wachenfeld.2016} \\
    TTCE & Time-to-closest-encounter & \cite{Eggert.2014} \\
    DCE & Distance-to-closest-encounter & \cite{Eggert.2014} \\
    CP & Collision probability & \cite{geisslingerconcept} \\
    HR & Harm and risk & \cite{GeißlerMaxRisk} \\
    BE & Break evaluation & \cite{Brannstrom.2008, Asljung2017} \\
    \bottomrule
\end{tabularx}
\label{tab:Metrics}
\end{table}
Further metrics can efficiently be integrated into the algorithm. While some of the presented metrics are well-established in the literature, we give particular attention to calculating the Harm and Risk (HR) and Break Evaluation (BE) metrics. The HR metric quantifies the potential harm and risk \cite{geisslingerconcept} associated with a trajectory by assessing the harm $H$ and probability $p$ of possible collisions with PAs, as shown in \cref{eq:trajrisk}. 
\begin{equation}
	R(\xi) = \mathrm{max}(p(\xi)H(\xi))
\label{eq:trajrisk}
\end{equation} 
The harm score is crucial for evaluating risks to VRUs, with collisions involving unprotected road users contributing more significantly to harm than those with protected users. Therefore, the module utilizes a harm model~\cite{geisslingerconcept} that quantifies harm based on the Abbreviated Injury Scale (AIS)~\cite{gennarelli2006ais}. It evaluates the relative velocity and angle of collision to estimate the probability of severe injury. This implicates computing the likelihood of an incident resulting in injuries classified as severity class 3 or higher (MAIS3+). The model outputs this probability on a normalized scale from $0$ to $1$. \\
The BE metric first calculates the minimum constant required acceleration $a_\mathrm{min, req}$ \cite{Brannstrom.2008} leveraging DCE. The required constant deceleration is the minimum deceleration to avoid a collision and is determined by an iterative approach. This value is then normalized against the maximum deceleration capability of the vehicle $a_\mathrm{veh, max}$ to infer the brake threat number (BTN), quantifying the required braking action \cite{Asljung2017}.
\begin{equation}
    \text{BTN} = \frac{a_\mathrm{min, req}}{a_\mathrm{veh, max}}
\end{equation}
The BE evaluates the efficacy of potential braking maneuvers, measuring the vehicle's ability to decelerate safely under current trajectory plans.

\subsection{Trajectory Safety Assessment}
\label{subsec:SafetyCheck}
In the final safety assessment of a trajectory, we compare the calculated metrics against their respective maximum thresholds. To illustrate, consider a trajectory $\xi \in \mathcal{T}$, where $R(\xi)$, $H(\xi)$, and $p(\xi)$ represent its risk, harm, and collision probability. A trajectory is valid if it adheres to the conditions:
\begin{equation}
R(\xi) < R_{\mathrm{max}} \land 
H(\xi) < H_{\mathrm{max}} \land 
p(\xi) < p_{\mathrm{max}}
\end{equation}
with $R_{\mathrm{max}}$, $H_{\mathrm{max}}$, and $p_{\mathrm{max}}$ being the set maximums for each metric. Should $\xi$ exceed any of these thresholds, it is classified as invalid.
Extending this principle to a more general form, let $M_i(\xi)$ denote a set of evaluated metrics for $\xi$, and $M_{i,\mathrm{max}}$ their corresponding maximum thresholds. The trajectory’s validity $v_{\xi}$ is then universally determined by:
\begin{equation}
v_{\xi} = 
\begin{cases}
\text{valid}, & \text{if } \forall i, M_i(\xi) < M_{i,\mathrm{max}} \\
\text{invalid}, & \text{otherwise}.
\end{cases}
\end{equation}
This condition considers all metrics, ensuring a trajectory is valid only if it meets all safety thresholds. The final evaluation is passed to the planner, which requests the trajectory safety check. Individual metric values are also provided to enable further assessments on the planner's side.

%!TeX spellcheck = en_US
% !TeX root = ../main.tex
\section{Results \& Analysis}
\label{sec:results}
%Our research aims to enhance the safety of autonomous driving by showcasing how motion planning, integrated with our proposed module, can effectively address complex situations.

\subsection{Simulation Setup and Scenarios}
\label{subsec:Integration}
\begin{figure}[!ht]
    \centering
    \resizebox{0.36\textwidth}{!}{\tikzset{every picture/.style={line width=0.75pt}} %set default line width to 0.75pt        

\begin{tikzpicture}[x=1pt,y=1pt,yscale=-1,xscale=1][trim axis left, trim axis right]
%uncomment if require: \path (0,294); %set diagram left start at 0, and has height of 294

%Shape: Trapezoid [id:dp052068149850935774] 
\draw  [draw opacity=0][fill={rgb, 255:red, 204; green, 204; blue, 204 }  ,fill opacity=1 ] (306,160) -- (317.23,190) -- (382.77,190) -- (394,160) -- cycle ;
%Shape: Trapezoid [id:dp3953650308739406] 
\draw   (296,90) -- (306.2,120) -- (393.8,120) -- (404,90) -- cycle ;
%Straight Lines [id:da7422112838561169] 
\draw    (276,94) -- (276,80) -- (350,80) -- (350,87) ;
\draw [shift={(350,90)}, rotate = 270] [fill={rgb, 255:red, 0; green, 0; blue, 0 }  ][line width=0.08]  [draw opacity=0] (6.25,-3) -- (0,0) -- (6.25,3) -- cycle    ;
%Shape: Rectangle [id:dp045416985795423415] 
\draw   (408,170) -- (408,110) -- (440,110) -- (440,170) -- cycle ;
%Straight Lines [id:da1308045562221617] 
\draw    (350,120) -- (350,127) ;
\draw [shift={(350,130)}, rotate = 270] [fill={rgb, 255:red, 0; green, 0; blue, 0 }  ][line width=0.08]  [draw opacity=0] (6.25,-3) -- (0,0) -- (6.25,3) -- cycle    ;
%Shape: Trapezoid [id:dp7299655845789486] 
\draw   (306,160) -- (317.74,190) -- (382.26,190) -- (394,160) -- cycle ;
%Shape: Rectangle [id:dp8962480199200232] 
\draw   (306,130) -- (394,130) -- (394,150) -- (306,150) -- cycle ;
%Straight Lines [id:da1777217338857252] 
\draw    (350,150) -- (350,157) ;
\draw [shift={(350,160)}, rotate = 270] [fill={rgb, 255:red, 0; green, 0; blue, 0 }  ][line width=0.08]  [draw opacity=0] (6.25,-3) -- (0,0) -- (6.25,3) -- cycle    ;
%Shape: Rectangle [id:dp5597674064139805] 
\draw   (266,184) -- (266,94) -- (286,94) -- (286,184) -- cycle ;
%Straight Lines [id:da4097930424898186] 
\draw    (350,190) -- (350,196) -- (424,196) -- (424,173) ;
\draw [shift={(424,170)}, rotate = 90] [fill={rgb, 255:red, 0; green, 0; blue, 0 }  ][line width=0.08]  [draw opacity=0] (6.25,-3) -- (0,0) -- (6.25,3) -- cycle    ;

% Text Node
\draw (270.5,176) node [anchor=north west][inner sep=0.75pt]  [rotate=-270] [align=left] {{\small Trajectory samples}};
% Text Node
\draw (350,140) node [anchor=center][inner sep=0.75pt]   [align=left] {{\small Optimal trajectory}};
% Text Node
\draw (350,169) node [anchor=center][inner sep=0.75pt]  [color={rgb, 255:red, 0; green, 0; blue, 0 }  ,opacity=1 ] [align=left] {{\small Occlusion}};
% Text Node
\draw (350,98) node [anchor=center][inner sep=0.75pt]   [align=left] {{\small Baseline planner}};
% Text Node
\draw (350,110) node [anchor=center][inner sep=0.75pt]   [align=left] {{\small trajectory evaluation}};
% Text Node
\draw (350,181) node [anchor=center][inner sep=0.75pt]  [color={rgb, 255:red, 0; green, 0; blue, 0 }  ,opacity=1 ] [align=left] {{\small evaluation}};
% Text Node
\draw (412,165) node [anchor=north west][inner sep=0.75pt]  [rotate=-270] [align=left] {{\small Optimal and }};
% Text Node
\draw (425,168) node [anchor=north west][inner sep=0.75pt]  [rotate=-270] [align=left] {{\small safe trajectory}};

\end{tikzpicture}}
    \caption{Enhanced evaluation funnel for occlusion-aware planning.}
    \label{fig:evaluation-funnel}
\end{figure}
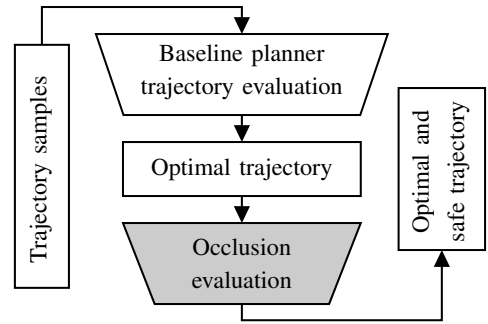

In this section, the proposed methodology is investigated using the CommonRoad framework \cite{commonroad}. For evaluating our module, we used an open-source trajectory planning algorithm that employs a Frenet coordinate system to generate multiple trajectory samples along a reference path $\Gamma$. These sampled trajectories then undergo checks for kinematic and dynamic feasibility, discarding any that are infeasible. The remaining trajectories are evaluated and ranked by weighted cost functions. The trajectory with the lowest cost is selected as optimal. This optimal trajectory is subject to a final collision check against static and quasi-static obstacles. If it is collision-free, it is used. Otherwise, the next trajectory that does not result in a collision is chosen. At this point, the FRENETIX-Occlusion module is integrated to expand the planner's evaluation funnel. \cref{fig:evaluation-funnel} shows the enhanced evaluation pipeline.

Our analysis employs four critical real-world scenarios adapted from the CommonRoad scenario database\footnote{{https://commonroad.in.tum.de/scenarios}} to replicate challenging situations. Our test scenarios are centered on turning maneuvers and intersections. This focus is informed by data showing a higher frequency of accidents in such situations, with a notable prevalence of personal injuries occurring in urban settings \cite{Morris.2021, NHTSADemographicTrends., Olszewski.2019, Puller.2021, Verkehrsunfaelle2022}. Scenarios with parked vehicles along the roadside are also evaluated. We create diverse possibilities by adjusting mission objectives and the states of obstacles. The baseline planning algorithm's settings are fixed from the outset and remain unchanged throughout the simulation. %This decision is critical as the results depend highly on these chosen wights. However, we vary the metric thresholds in the trajectory safety assessment to demonstrate the vehicle's behavior dependency on the selected thresholds. 
This approach ensures that our results reflect the performance of our algorithm, free from biases introduced by mid-simulation adjustments. The fixed cost weights for the baseline planner are depicted in \cref{tab:cost_weights}.
\begin{table}[h]
\centering
\caption{Untuned baseline planner cost weights.}
\begin{tabularx}{0.6\linewidth}{p{3cm} c}
    \toprule
    Cost function & Cost weights \\
    \midrule
    Lateral jerk    & \SI{1.0}{}           \\ 
    Longitudinal jerk    & \SI{1.0}{}   \\
    Distance to reference path    & \SI{3.0}{}    \\  
    Velocity    & \SI{0.1}{}  \\ 
    Distance to obstacles    & \SI{0.1}{}  \\ 
    Collision probability    & \SI{200}{}   \\
    \bottomrule
    \end{tabularx}
\label{tab:cost_weights}
\end{table}

\subsection{Simulation Results}
\begin{figure}[!ht]
    \centering
    \hspace{1mm}
    %\vspace{-1cm}
    \begin{tikzpicture}[font=\scriptsize]
        % Ego vehicle
        \node[inner sep=0pt] at (0.6,0) {\includegraphics[height=2.5mm]{figures/ego.png}};
        \node[align=left, anchor=west] at (0.9,0) {ego \\ vehicle};
    
        % Obstacle
        \node[inner sep=0pt] at (2.3,0) {\includegraphics[height=2.5mm]{figures/dyn_obstacle.png}};
        \node[align=left, anchor=west] at (2.6,0) {dynamic \\ obstacle};
    
         % trajectory
        \draw[thick, black, line width=1pt] (3.8,0) -- (4.1,0);
        \node[align=left, anchor=west] at (4.2, 0) {trajectory};

        % prediction
        \node[inner sep=0pt] at (5.6,0) {\phantomPrediction};
        \node[align=left, anchor=west] at (5.8,0) {PA \\ prediction};

        % PA
        \node[inner sep=0pt] at (7.3,0) {\phantomBike};
        \node[inner sep=0pt] at (7.8,0) {\phantomPedestrian};
        \node[align=left, anchor=west] at (8.0,0) {PAs};
        
    \end{tikzpicture}
    \includegraphics[width=0.7\linewidth, trim={1.2cm 1.5cm 2.5cm 2.0cm},clip]{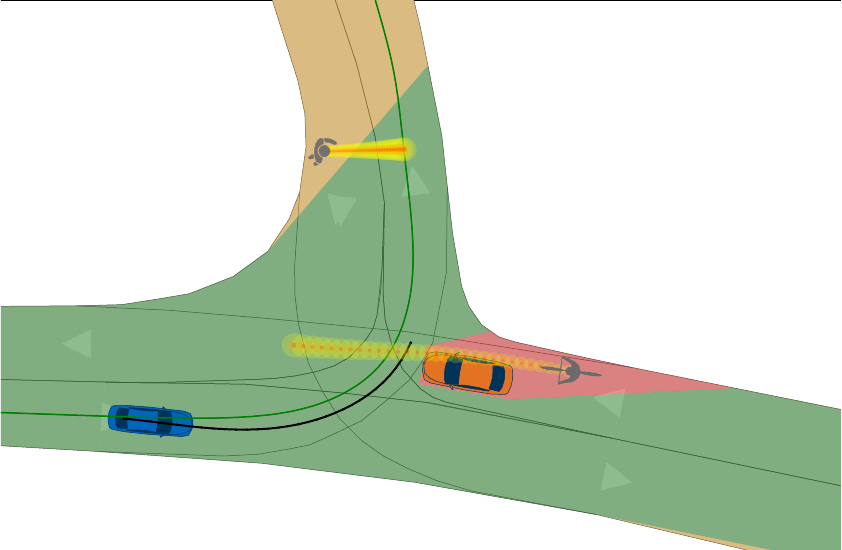}
    \caption{Scenario 1: A representation of an intersection where an autonomous vehicle navigates visible and occluded zones, highlighting the importance of anticipating PAs' movements for safe trajectory planning.}
    \label{fig:visualization_phantom_scenario}
\end{figure}

\textbf{Safer driving behavior through occlusion aware planning:} \cref{fig:visualization_phantom_scenario} presents Scenario~1, an exemplary intersection where the AV is tasked with making a left turn. It encounters a section of the road occluded by a dynamic obstacle and another area obscured by the street geometry.

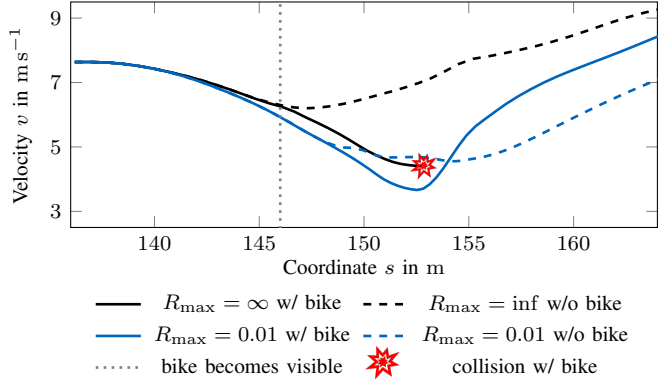
\begin{figure}[!ht]
    \centering
    \begin{tikzpicture}[font=\footnotesize]

\definecolor{darkcyan0101189}{RGB}{0,101,189}
\definecolor{darkgray176}{RGB}{176,176,176}
\definecolor{darkred140015}{RGB}{140,0,15}
\definecolor{gray127}{RGB}{127,127,127}
\definecolor{lightgray204}{RGB}{204,204,204}
\definecolor{midnightblue05189}{RGB}{0,51,89}

\begin{axis}[
/pgf/number format/.cd,
1000 sep={},
height=3.0cm,
width=7.8cm,
legend style={
	at={(0.5,-0.25)}, 
	anchor=north,
	legend columns=2,
	cells={anchor=center},
	draw=none,
	column sep=0.25em,
	row sep=0.1em,
},
scale only axis,
xlabel={Coordinate $s$ in \si{\meter}},
ylabel={Velocity $v$ in \si{\meter\per\second}},
x label style={at={(0.5,-0.1)},anchor=north},
y label style={at={(-0.05,.5)},anchor=south},
xmin=136, xmax=164,
ymin=2.5, ymax=9.5,
ytick = {3, 5, ..., 9},
%ymajorgrids=true
]

\addplot [line width=1pt, black]
table {%
136.195 7.63477
136.959 7.63573
137.722 7.62196
138.482 7.57979
139.237 7.5141
139.984 7.42677
140.72 7.32339
141.445 7.21081
142.156 7.0853
142.851 6.9449
143.53 6.79755
144.189 6.6435
144.83 6.49229
145.452 6.35868
146.058 6.24886
146.651 6.10166
147.224 5.91673
147.774 5.74899
148.302 5.57952
148.805 5.40477
149.285 5.22483
149.742 5.04
150.175 4.87292
150.59 4.73882
150.99 4.62514
151.378 4.53321
151.756 4.47237
152.128 4.43532
152.493 4.41226
152.853 4.40145
};
\addlegendentry{$R_\mathrm{max} = \infty$ w/ bike}
\addplot [line width=1pt, black, dashed]
table {%
136.195 7.63477
136.959 7.63573
137.722 7.62196
138.482 7.57979
139.237 7.5141
139.984 7.42677
140.72 7.32339
141.445 7.21081
142.156 7.0853
142.851 6.9449
143.53 6.79755
144.189 6.6435
144.83 6.50165
145.454 6.38549
146.064 6.29154
146.664 6.22629
147.258 6.20011
147.846 6.21931
148.433 6.26725
149.018 6.33713
149.604 6.42506
150.192 6.52531
150.783 6.63086
151.38 6.73446
151.983 6.8509
152.594 6.98111
153.216 7.13688
153.855 7.34517
154.516 7.57432
155.203 7.73496
155.915 7.80927
156.653 7.90325
157.415 8.00965
158.2 8.12807
159.005 8.26989
159.831 8.43115
160.679 8.61398
161.549 8.81085
162.439 8.99952
163.347 9.16264
164.271 9.30209
165.207 9.42107
166.154 9.52114
167.111 9.60271
168.074 9.66819
169.044 9.71647
170.016 9.74541
170.991 9.75643
171.965 9.74589
172.937 9.7189
173.907 9.68396
};
\addlegendentry{$R_\mathrm{max} = \inf$ w/o bike}
\addplot [line width=1pt, Bluelight]
table {%
136.195 7.63477
136.959 7.63573
137.722 7.62196
138.482 7.57979
139.237 7.5141
139.984 7.42677
140.72 7.31634
141.443 7.18354
142.151 7.03183
142.842 6.86551
143.514 6.69227
144.166 6.51486
144.798 6.32924
145.407 6.13351
145.995 5.92734
146.558 5.71577
147.098 5.51288
147.614 5.33011
148.108 5.15742
148.579 4.98811
149.028 4.82034
149.456 4.65229
149.862 4.48233
150.247 4.31048
150.612 4.13838
150.957 3.98859
151.289 3.87773
151.61 3.79425
151.923 3.73045
152.229 3.68742
152.531 3.66264
152.83 3.70329
153.134 3.84
153.447 4.04812
153.776 4.32679
154.124 4.65981
154.495 5.0175
154.891 5.35911
155.313 5.63613
155.765 5.86128
156.249 6.09011
156.769 6.33097
157.328 6.56009
157.925 6.77993
158.561 6.99274
159.233 7.1921
159.936 7.37877
160.668 7.56335
161.426 7.75157
162.209 7.94811
163.014 8.15933
163.842 8.38138
164.693 8.60838
165.565 8.81965
166.456 8.99887
167.363 9.14054
168.282 9.25287
169.213 9.34974
170.152 9.43465
171.099 9.50875
172.052 9.56452
173.008 9.60242
173.969 9.62527
};
\addlegendentry{$R_\mathrm{max} = 0.01$ w/ bike}

\addplot [line width=1pt, Bluelight, dashed]
table {%
136.195 7.63477
136.959 7.63573
137.722 7.62196
138.482 7.57979
139.237 7.5141
139.984 7.42677
140.72 7.31634
141.443 7.18354
142.151 7.03183
142.842 6.86551
143.514 6.69227
144.166 6.51486
144.798 6.32924
145.407 6.13351
145.995 5.93164
146.559 5.72377
147.1 5.51573
147.616 5.32442
148.112 5.16299
148.589 5.04743
149.055 4.98191
149.513 4.96615
149.965 4.89509
150.403 4.77285
150.829 4.69994
151.248 4.67315
151.665 4.67649
152.08 4.68443
152.494 4.68474
152.906 4.67208
153.314 4.64074
153.719 4.58907
154.12 4.55093
154.521 4.56225
154.929 4.60358
155.347 4.64808
155.777 4.70234
156.221 4.77806
156.68 4.87904
157.156 5.00216
157.65 5.14432
158.163 5.30502
158.695 5.47877
159.248 5.66022
159.821 5.84705
160.415 6.03806
161.028 6.23165
161.662 6.42622
162.315 6.62031
162.987 6.81241
163.678 7.00093
164.388 7.18593
165.116 7.36726
165.861 7.54469
166.624 7.70923
167.402 7.85236
168.194 7.97733
168.997 8.08692
169.811 8.1828
170.634 8.26682
171.464 8.34037
172.302 8.40388
173.145 8.45834
173.993 8.50467
174.845 8.54381
175.701 8.57707
176.56 8.60488
177.422 8.62808
178.286 8.64721
179.151 8.66265
180.018 8.67469
180.886 8.68403
181.755 8.69103
182.624 8.69599
183.494 8.69931
184.364 8.70113
185.234 8.70157
186.104 8.7008
186.974 8.69897
187.844 8.69622
188.713 8.69268
};
\addlegendentry{$R_\mathrm{max} = 0.01$ w/o bike}

\addplot[draw=Grey, line width=1pt, dotted] coordinates {(146,0) (146,13)};
\addlegendentry{bike becomes visible}

\node[inner sep=0pt] at (152.853, 4.40145) {\crashIconsmall};

\addplot [only marks, mark=x, mark size=0pt, mark options={color=white, thick}] coordinates {(140, 5)};
\addlegendentry{collision w/ bike}

\end{axis}

\node[inner sep=0pt] at (4.15, -1.8) {\crashIcon};
\node[inner sep=0pt] at (10.15, -1.7) {};

\end{tikzpicture}
    \vspace{-2em}
    \caption{Velocity profiles for Scenario 1 across four simulation runs, illustrating the autonomous vehicle's response to varying risk thresholds.}
    \label{fig:scenario1_s_v}
\end{figure}

\noindent Additionally, the illustration shows PAs and their predicted movements, which are utilized to assess the safety of the ego vehicle's planned trajectory. Scenario 1 was evaluated through four simulation runs designed to determine the impact of different settings on the planner's performance. Two simulations were conducted with an unrestricted maximum risk threshold $R_\mathrm{max}$, and two with a limited $R_\mathrm{max}$. Each pair of simulations included one iteration with a real cyclist positioned in the occluded area behind the car and another without a real cyclist. The resulting velocity profiles from these simulations are illustrated in \cref{fig:scenario1_s_v}, providing insights into how the presence of a real cyclist and the adjustment of the risk threshold influence the autonomous vehicle's trajectory.

\begin{figure}[!t]
    \centering
    \input{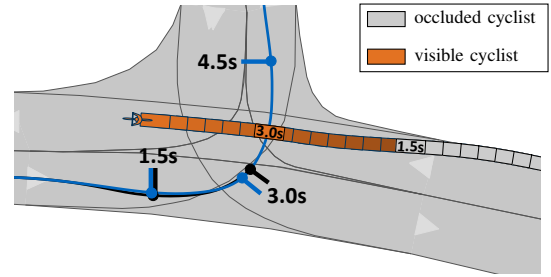}
    \caption{Final trajectories for Scenario 1 with a cyclist emerging behind a car. The black trajectory collides due to the higher velocity.}
    \label{fig:scenario1_x_y}
\end{figure}

\noindent The figure shows that the vehicle with the limited $R_\mathrm{max}$ decelerates earlier in both settings. When $R_\mathrm{max}$ is unrestricted, braking commences only once the cyclist becomes visible. However, since the speed is then too high, the vehicle cannot decelerate fast enough, ultimately leading to a collision. This circumstance is further illustrated in \cref{fig:scenario1_x_y}, depicting the vehicle's trajectory and positions at selected timesteps. 
The earlier deceleration with a limited $R_\mathrm{max}$ results in the vehicle keeping a greater distance from the cyclist, thereby avoiding a collision. Subsequently, the simulation can continue, with the car accelerating back to the target speed. To further evaluate the planner's response under varying conditions and metrics, two additional test scenarios were simulated with different levels of harm limitation $H_\mathrm{max}$.

\begin{figure*}[!t]
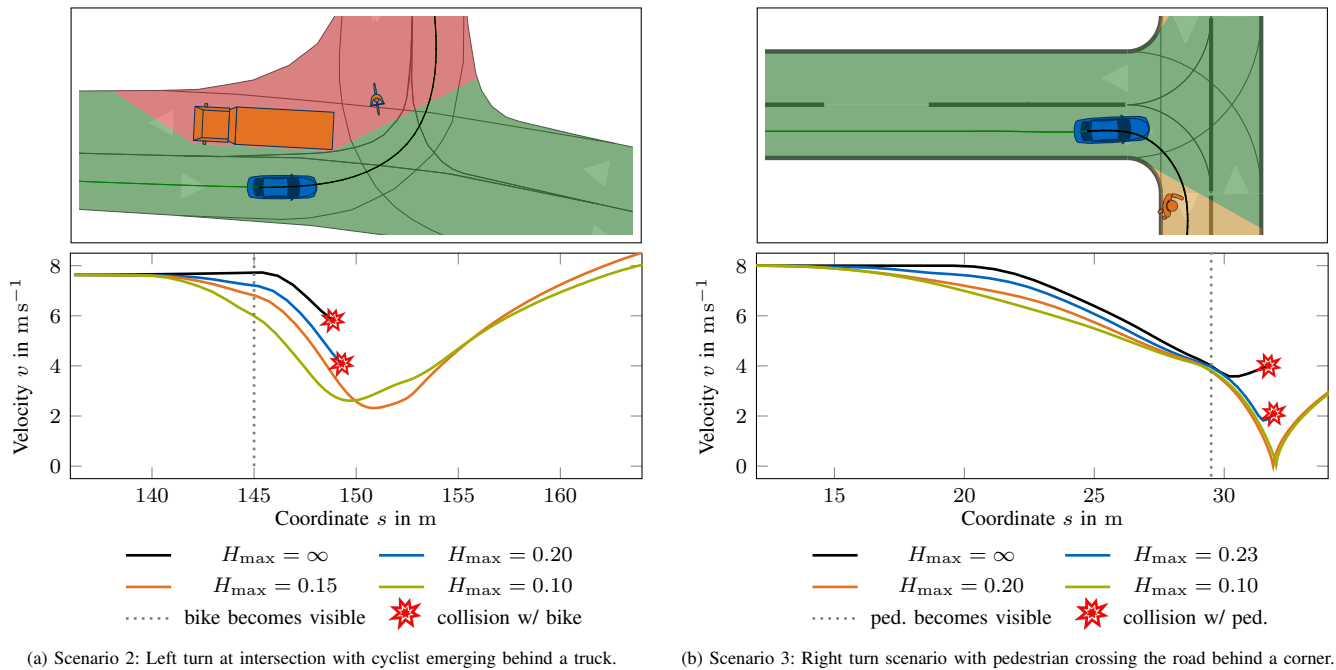

    \centering
    \subfloat[][Scenario 2: Left turn at intersection with cyclist emerging behind a truck.]{\label{fig:scenario2_s_v} \input{figures/results/scenario2_s_v}} \quad
    \subfloat[][Scenario 3: Right turn scenario with pedestrian crossing the road behind a corner.]{\label{fig:scenario3_s_v} \input{figures/results/scenario3_s_v}}
    \caption{Scenario visualization and velocity profiles for Scenario~2 and Scenario~3 across four simulation runs with different harm thresholds $H_\mathrm{max}$. The lower the harm threshold is set, the earlier and more significantly the vehicle decelerates, enhancing its ability to avoid potential collisions.}%
    \label{fig:scenario2_scenario3}%
\end{figure*}

Scenario 2 displays the same intersection but now with an oncoming truck that obstructs the view into the intersection (\cref{fig:scenario2_s_v}). Furthermore, in Scenario 3, a right-turn maneuver (\cref{fig:scenario3_s_v}) was simulated. In both scenarios, vulnerable road users (VRUs) are present in the occluded areas. The dotted line in the velocity profiles marks the moment when the cyclist in Scenario 2 and the pedestrian in Scenario 3 become visible to the vehicle's sensors. Using the baseline planner, risks arising from occluded areas are not accounted for. This leads to the selection of a higher speed, resulting in a collision due to insufficient braking time. Without any harm restriction in Scenario 2, it is evident that the braking process is only initiated once the cyclist becomes visible. When maximum acceptable harm values $H_\mathrm{max}$ are specified, the velocity profiles show that the vehicle decelerates accordingly. The more restrictive the threshold, the earlier and more significantly the vehicle initiates braking. After passing the critical blind spots, the ego vehicle accelerates to its desired speed. The specified thresholds thus only result in deceleration when passing critical areas; otherwise, the vehicle operates without restrictions.
The outcomes of additional simulation runs are presented in \cref{tab:scenario_evaluation}, where we expanded the evaluation to include a scenario with parked vehicles along the road (Scenario 4). In each scenario, we assessed the possibility of a collision with PAs that may emerge from occluded areas. Different metrics were evaluated to determine the vehicle's response to these scenarios.
\begin{table}[!ht]
\centering
\caption{Evaluation of scenarios 1 and 4 using exemplary metrics.}
\begin{tabularx}{0.88\linewidth}{c c c c}
    \toprule
    Scenario & Metric $M_\mathrm{i}$ & min. velocity $v_\mathrm{min}$ & collision \\
    \midrule
    \multirow{5}{*}{1} & baseline, no limits & \SI{4.396}{\meter\per\second} & \checkmark \\
    & $\text{BTN}_\mathrm{max} = 0.1$ & \SI{3.053}{\meter\per\second} & $\times$ \\
    & $\text{BTN}_\mathrm{max} = 0.2$ & \SI{2.225}{\meter\per\second} & \checkmark \\
    & $\text{DCE}_\mathrm{min} = \SI{1.0}{\meter}$ & \SI{4.133}{\meter\per\second} & \checkmark \\
    & $\text{DCE}_\mathrm{min} = \SI{2.0}{\meter}$ & \SI{3.588}{\meter\per\second} & $\times$ \\
    \midrule
    \multirow{5}{*}{4} & baseline, no limits & \SI{4.874}{\meter\per\second} & \checkmark \\
    & $\text{BTN}_\mathrm{max} = 0.3$ & \SI{3.019}{\meter\per\second} & \checkmark \\
    & $\text{BTN}_\mathrm{max} = 0.2$ & \SI{1.576}{\meter\per\second} & $\times$ \\
    & $H_\mathrm{max} = 0.2$ & \SI{3.757}{\meter\per\second} & \checkmark \\
    & $H_\mathrm{max} = 0.1$ & \SI{0.743}{\meter\per\second} & $\times$ \\
    \bottomrule
    \end{tabularx}
\label{tab:scenario_evaluation}
\end{table}

It is observed that stricter metric limits lead to lower minimum velocities, which can prevent collisions. Conversely, when no limits are applied, as seen with the baseline planner, or when the limits are less strict, the algorithm permits higher velocities, which may result in collisions. The table also illustrates that similar results can be achieved by applying different criticality metrics. A detailed analysis of Scenario 4 is omitted as it has already been examined in \cite{TrauthOJITS}.

\textbf{Runtime analysis:}
For the evaluation of the computation times of our module, the experiments were conducted on a standard laptop powered by a 12th Gen Intel\textsuperscript{\textregistered} Core\texttrademark{} i7-1270P processor with 16 threads and 32 GB of RAM. The runtime analysis for metric evaluation is detailed in \cref{tab:runtime_metrics}. Each metric is assessed for a given trajectory against an average of two PAs, showcasing the computational efficiency of the system. The TTC, WTTC and TTCE metrics are derived from the DCE metric. Hence, the TTC is exemplified in the table to represent this category. The processing times for these metrics generally range from approximately \SIrange{0.5}{4}{\milli\second} per trajectory. However, the BE metric often requires longer computation times, as it involves an iterative process to calculate the minimal necessary deceleration, incorporating numerous collision checks.
\begin{table}[!ht]
\centering
\caption{Runtime for different metrics per trajectory in \SI{}{\milli\second}.}
\begin{tabularx}{0.85\linewidth}{l c c c c c}
    \toprule
    & CP & DCE & TTC & HR & BE \\
    \midrule
    min & 0.0777 & 2.4519 & 0.0484 & 0.0911 & 0.0017 \\
    $P_\mathrm{25}$ & 0.7219 & 4.1432 & 0.0739 & 0.1719 & 14.7536  \\
    median & 0.7939 & 4.3278 & 0.0803 & 0.1805 & 15.5444  \\
    $P_\mathrm{75}$ & 0.8807 & 4.528 & 0.0858 & 0.1945 & 23.2432  \\
    max & 1.8289 & 10.5934 & 0.1483 & 0.4063 & 40.7581  \\
    \bottomrule
    \end{tabularx}
\label{tab:runtime_metrics}
\end{table}

In \cref{tab:runtime_functions}, the runtimes of individual functions within our module are analyzed per timestep. The sensor model (SM), spawn point prediction (SPP), and pedestrian PA prediction (PPA) are collectively computed typically in less than \SI{25}{\milli \second} per timestep, demonstrating swift performance. In contrast, the prediction for vehicle PAs (VPA) takes longer due to the complex computation of potential paths, initialization of the curvilinear coordinate system, and trajectory calculation.
\begin{table}[!ht]
\centering
\caption{Runtime for different functions per timestep in \SI{}{\milli\second}.}
\begin{tabularx}{0.75\linewidth}{l c c c c}
    \toprule
    & SM & SPP & VPA pred. & PPA pred. \\
    \midrule
    min & 8.769 & 3.360 & 49.442 & 0.194 \\
    $P_\mathrm{25}$ & 10.004 & 4.509 & 98.924 & 0.344 \\
    median & 16.190 & 7.081 & 109.969 & 0.389 \\
    $P_\mathrm{75}$ & 16.974 & 7.672 & 161.096 & 0.420 \\
    max & 27.171 & 7.838 & 183.301 & 0.568 \\
    \bottomrule
    \end{tabularx}
\label{tab:runtime_functions}
\end{table}

%!TeX spellcheck = en_US
% !TeX root = ../main.tex
\section{Discussion}
\label{sec:discussion}
Our simulations show that the FRENETIX-Occlusion module can improve safer AV behavior near blind spots caused by sensor occlusions. We could show a safer driving behavior by selecting appropriate thresholds for distinguished criticality metrics and incorporating them into the motion planning algorithm. This is mainly caused by the vehicle reducing speed in critical situations to prevent collisions. The modular structure of our module allows fast adaptation and integration of new metrics. Nonetheless, the presented criticality thresholds are not universally applicable across all traffic situations. Therefore, future research should aim for a situation-adapted determination of these thresholds. The computational demands, particularly for predicting vehicle movements, are also considerable. If many trajectories are discarded as part of the safety assessment, the computing load increases due to repeated metric calculations. Efficiency gains would lead to further performance improvements.

%To address this, parallel computing techniques or offloading computations to a C++ environment could offer viable solutions to enhance processing times. Even though spawn points are fundamentally identified in critical occluded areas, the current approach considers only instantaneous information and not the temporal progression, potentially marking points as critical that were visible moments before.

\section{Conclusion \& Outlook}
\label{sec:conclusion}
%Our study delves into the complexities of autonomous driving in dynamic environments, mainly focusing on the challenges presented by occluded areas. We provide an open-source module for conducting trajectory safety checks in occluded scenarios, which can be integrated into existing trajectory planning algorithms.
This paper presents our FRENETIX-Occlusion module, which enhances AV safety in occluded urban environments. Our approach is able to identify potential occluded traffic areas and model traffic participants as phantom agents. The module predicts their possible movements, leading to the calculation of various criticality metrics. When this occlusion-aware module is combined with a trajectory planning algorithm, the calculations allow for adjusting driving behavior in complex scenarios. The results from simulating real-world scenarios using the occlusion-aware module showed remarkable findings. The results demonstrate our module's capability to modify vehicle behavior in occluded scenarios, ensuring higher AV safety. This module is released open-source, laying the groundwork for continued development and refinement within the community. Looking ahead, there is scope for further enhancement. One potential avenue is incorporating temporal tracking, similar to the techniques outlined in \cite{Sanchez.2022}. Furthermore, offloading computations to a C++ environment could offer viable solutions to enhance processing times.


\end{document}